%% file: main.tex
\documentclass[man, floatsintext] {apa7}
\usepackage{graphicx}  
\usepackage{caption}   
\usepackage{float}     
\usepackage{lipsum}
\usepackage{amsfonts}       
\usepackage{amsmath,amsfonts, amssymb}
\usepackage{makecell}
\usepackage{enumitem}
\DeclareMathOperator*{\argmax}{arg\,max}
\newcommand{\probP}{\text{I\kern-0.15em P}}
\usepackage{xr} 
\usepackage{subfiles} 

\usepackage[american]{babel}
\usepackage{csquotes}
\usepackage[style=apa,sortcites=true,sorting=nyt,backend=biber]{biblatex}
\DeclareLanguageMapping{american}{american-apa}
\title{Latent Variable Sequence Identification for Cognitive Models with Neural Network Estimators}
\shorttitle{LaseNet}

\authorsnames[1,2,1,{1,2}]{Ti-Fen Pan, Jing-Jing Li, Bill  Thompson, Anne GE Collins}

\authorsaffiliations{{Department of Psychology, University of California, Berkeley},{Helen Wills Neuroscience Institute, University of California, Berkeley}}

\leftheader{Weiss}
\abstract{Extracting time-varying latent variables from computational cognitive models plays a key role in uncovering the dynamic cognitive processes that drive behaviors.  
However, existing methods are limited to inferring auxiliary latent variables in a relatively narrow class of cognitive models. For example, a broad class of relevant cognitive models with analytically intractable likelihood is currently out of reach from standard techniques, based on Maximum a Posteriori parameter estimation. Here, we present a simulation-based approach that leverages recurrent neural networks to map experimental data directly to the targeted latent variable space. We show that our approach achieves competitive performance in inferring latent variable sequences in both likelihood-tractable and intractable models. Furthermore, the approach is practical to standard-size, individual participants' data, generalizable across different computational models, and adaptable for continuous and discrete latent spaces. We then demonstrate its applicability in real-world datasets. Our work underscores that combining recurrent neural networks and simulated data to identify auxiliary latent variable sequences broadens the scope of cognitive models researchers can explore, enabling testing a wider range of theories.}

\keywords{computational cognitive models, intractable likelihood, artificial neural networks, auxiliary latent variables}

\authornote{
  Corresponding author: \addORCIDlink{Ti-Fen Pan}{0000-0002-2640-7664}, Department of Psychology, University of California, Berkeley, 2121 Berkeley Way, 94704.  E-mail: tfpan@berkeley.edu}

\begin{document}
\maketitle
Researchers often use auxiliary time-varying latent variables from computational cognitive models to capture the dynamics of cognitive processes. These variables provide trial-by-trial predictors of internal states, enabling investigation into underlying mechanisms and individual differences \parencite{katahira2021revisiting}. For instance, psychologists apply state-space models (e.g., hidden Markov models) to infer time-varying attentive states, offering better explanations for noisy behaviors \parencite{ashwood2022mice, li2024dynamic}.  In neuroscience, model-based analyses commonly link neural activity to latent variables derived from computational cognitive models, illuminating how the brain supports cognition  \parencite{cohen2017computational}.  Notably, reward prediction errors (RPE) extracted from a reinforcement learning (RL) model have been found to correlate with ventral striatum activity in human functional magnetic resonance imaging (fMRI) \parencite{o2007model} studies, as well as phasic activity of dopamine neurons in non-human animals \parencite{eshel2016dopamine}. 

The traditional method for extracting time-varying latent variables from experimental data consists of two steps \parencite{wilson2019ten}: first, identify the best-fitting model parameters. Second, infer the auxiliary latent variables by running the computational model off-policy over participants' experienced sequences of stimuli and actions with the best-fitting model parameters. However, both steps limit the type of models that can be considered. In the first step, best-fitting parameters may be difficult to obtain in a large subspace of relevant cognitive models. Indeed, researchers typically use likelihood-dependent methods like Maximum Likelihood Estimation (MLE) \parencite{myung2003tutorial}, Maximum a Posteriori (MAP) \parencite{cousineau2013improving} or hierarchical Bayesian modeling \parencite{baribault2023troubleshooting}. Nonetheless, these methods cannot be used for models with analytically intractable likelihood \parencite{lenzi2023neural}. In addition, auxiliary latent variables may be complex to infer even when best-fitting parameters are known, requiring custom design of statistical tools specific to a model and experiment \parencite{findling2021imprecise}. Due to these limitations, testing a broader class of theories is significantly slowed down by the need for researchers to develop customized statistical approaches that are not generalizable to broader computational models \parencite{escola2011hidden, ashwood2022mice}. 

Most computational models with intractable likelihoods can be simulated, making simulation-based inference (SBI) a powerful approach to bypass the likelihood computation \parencite{busetto2013approximate}. SBI methods, particularly when combined with artificial neural networks (ANNs), have proven effective in parameter recovery across various computational models. ANNs excel in handling high-dimensional data and enabling amortized inference after training. These neural-based SBI methods are primarily based on Bayesian inference, aiming to approximate likelihood (neural likelihood estimation, NLE) \parencite{boelts2022flexible} or the posterior (neural posterior estimation, NPE) \parencite{radev2020bayesflow} (see the recent review \cite{zammit2024neural}).  Recently, the approach consisting of training neural networks to map data to parameter point estimates has shown promising avenues in a variety of applications \parencite{lenzi2023neural, sainsbury2023neural}. \textcite{sainsbury2024likelihood} conceptualized this method by connecting it to classic Bayes estimators, which we refer to as neural point estimators here. 

However, the extraction of time-varying latent variables in likelihood intractable models is still under-explored \parencite{schumacher2023neural}. Existing SBI methods focus mainly on parameter recovery or model identification \parencite{gloeckler2024all, zammit2024neural}. Although SBI methods help recover the model parameters, approximating auxiliary latent variable sequences often proves challenging. This is especially true in models with sequential dependencies, where sampling methods like sequential Monte Carlo (SMC) \parencite{gordon1993novel, doucet2000sequential, samejima2003estimating} can be computationally expensive when reconstructing latent variables \parencite{ghosh2024sample}. 

Here, we propose an approach that aims to identify \textbf{La}tent variable \textbf{Se}quences with ANNs, which we call "LaseNet".  LaseNet is built on neural point estimators, given their success in applying to computational cognitive models with sequential dependencies and intractable likelihoods \parencite{rmus2023artificial}. We use simulated datasets to train an ANN, which learns a direct mapping between a sequence of observable variables (e.g., the participant's actions or received outcomes) and the targeted latent variable space (e.g., the participant's reward expectation or subjective rule choice). We begin by outlining the problem formalism and cognitive models in the following section, followed by a detailed introduction of the LaseNet in section \nameref{method}.  In section \nameref{benchmark}, we show that LaseNet infers time-varying latent variables that are close to synthetic ground truth for a variety of computational cognitive models and task environments.  We highlight LaseNet's real-world applicability in section \nameref{real-data}: using experimental data from a real mice dataset, our approach successfully infers both discrete and continuous latent variables compared to likelihood-dependent estimations. Finally, we discuss related work and the benefits and limitations of LaseNet.

\subsection{Problem formulation}
Suppose that there is a latent variable model that produces a time-series of observable variables $Y=(y_{1}, y_{2}...y_{T})$ and unobservable auxiliary latent variables $Z=(z_{1}, z_{2}...z_{T})$ given a set of model parameters $\theta$, where $T$ denotes the number of trials.  We can then describe the generation process for the time-varying latent variables as follows:
\begin{equation}
z_t\sim f(\mathbf{z_{t-1}}, \overline{y}_{t-1}, \theta_f),\; \; \; \; \: \; y_t\sim g(z_{t},  \theta_g)
\end{equation}
where $f$ and $g$ are both density functions parameterized by $\theta_f$ and $\theta_g$ respectively, $\overline{y}_{t-1}$ corresponds to the history of $Y$ up to the trial $t-1$. Our goal is to infer the unobservable latent variables $Z$ at each time point given variables $Y$. Therefore, our objective can be described as:
\begin{equation}
\widetilde{Z} =\argmax_{Z \in \text{K}}\, \, P(Z \mid \theta, Y)P(\theta \mid Y)
\end{equation}
where K denotes the possible values in the latent variable space, which can be either continuous $Z_{c}$ or discrete $Z_{d}$. $\theta$ includes $(\theta_f, \theta_g)$. Note that we do not assume the Markovian property in the latent variable model and will demonstrate how LaseNet can apply to cognitive models with and without the Markovian property.

\subsection{Task environments and computational cognitive models}
\label{preliminaries}
We illustrate our approach with representative examples of sequential learning and decision-making tasks, where biological agents (e.g., humans, mice) are assumed to be in a state (for example defined by an observable stimulus), select actions, and may observe outcomes. Examples of stimuli include specific images or abstract features characterized along a specific dimension, such as the orientation of a grating. Agents' choices typically correspond to pressing one of multiple keys or levers; subsequently, feedback outcomes are typically points for humans or water for mice. Computational cognitive models are simple algorithms that operate over few variables (typically $<10$); they instantiate specific hypotheses about the information flow and provide quantitative predictions about behavioral data. To illustrate our technique, we use three representative families of cognitive models as follows: 
\paragraph{Reinforcement learning (RL) models} RL cognitive models, such as delta rule or Q-learning \parencite{niv2012neural, sutton2018reinforcement, eckstein2022reinforcement} assume that an agent tracks the Q-value of actions, and uses these Q-values to inform action selection on each trial. After each trial's outcome, the model updates Q-values by first computing the reward prediction error (RPE), denoted by $\delta$, as the discrepancy between the expected and the observed values, and then adjusting the Q-value of the chosen action $a$ with RPE scaled by a learning rate $\alpha$ \parencite{sutton2018reinforcement}:
\begin{equation}
\label{eq:q-learning}
\begin{gathered}
\delta_t = r_t - Q_t(a_t) \\ 
Q_{t+1}(a_t)=Q_t(a_t) + \alpha\,\delta_t 
\end{gathered}
\end{equation}
One example of a time-varying latent variable of interest to model-based analyses \parencite{cohen2017computational} is Q-values ($Z_{c}$), which are inferred from observable rewards, stimuli, and actions ($Y$). This model is typically likelihood \emph{tractable}, providing acbaseline standard approaches. In addition, we explore a hierarchical extension of this model, where choices are governed by higher-level rules that are not directly observable, making the model’s likelihood \emph{intractable}. This allows us to investigate latent rule inference over discrete latent states ($Z_{d}$). 

\paragraph{Bayesian generative models} We consider a variant of hierarchical Bayesian generative models that incorporates noisy inference following Weber’s law \parencite{fechner1860elemente, findling2021imprecise}, positing that computational imprecision scales with the distance between posterior beliefs in consecutive trials ($t$ and $t+1$). This model is likelihood-intractable; however, a custom statistical approach was developed, providing a baseline for our approach. Our study aims to infer both the agents’ discrete latent state ($Z_{d}$) and the distance ($Z_{c}$) in posterior beliefs across trials given observable rewards, stimuli, and actions ($Y$).
\paragraph{Hidden Markov models (HMMs)} We use a cognitive model based on the Bernoulli generalized linear model (GLM) observations (GLM-HMM) \parencite{escola2011hidden, calhoun2019unsupervised, ashwood2022mice}.  Again, this model is likelihood-intractable, but a custom statistical approach provides a benchmark. Our example target output is the HMM hidden states ($Z_{d}$) given rewards/stimuli and actions ($Y$). See Appendix \ref{sup:ccm} for more detailed descriptions of the cognitive models.

\section{Method}\label{method}
\subsection{RNN-based Inference for Time-Varying Latent Variables}
\begin{figure}[h]
    \centering
    \includegraphics[width=\linewidth]{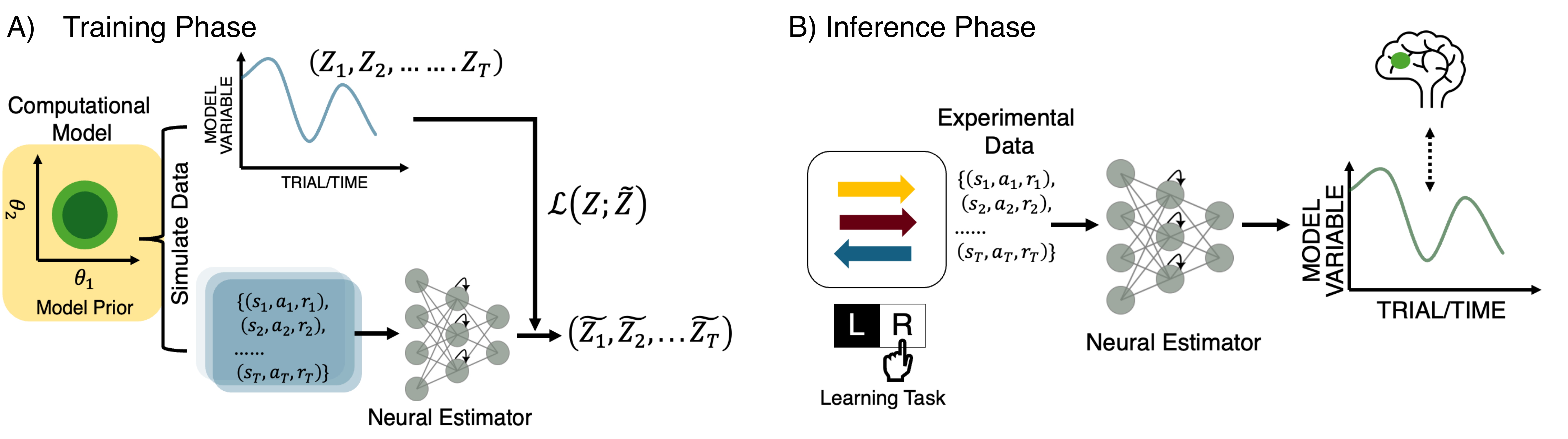}
    \caption{\textbf{Schematic of the LaseNet method}: (A) The network is trained with a simulated dataset to predict time-varying latent variables derived from a computational cognitive model. The input may include a time-series of simulated stimuli, actions, and rewards. Corresponding latent variables are used as training targets. (B) At inference time, the trained network predicts the latent variables for experimental data from biological agents where ground truth is unknown. The extracted latent variables are commonly used in relating behavior to brain signals.}
    
    \label{fig:approach}
\end{figure}

\begin{figure}[H]
    \centering
    \includegraphics[width=0.8\linewidth]{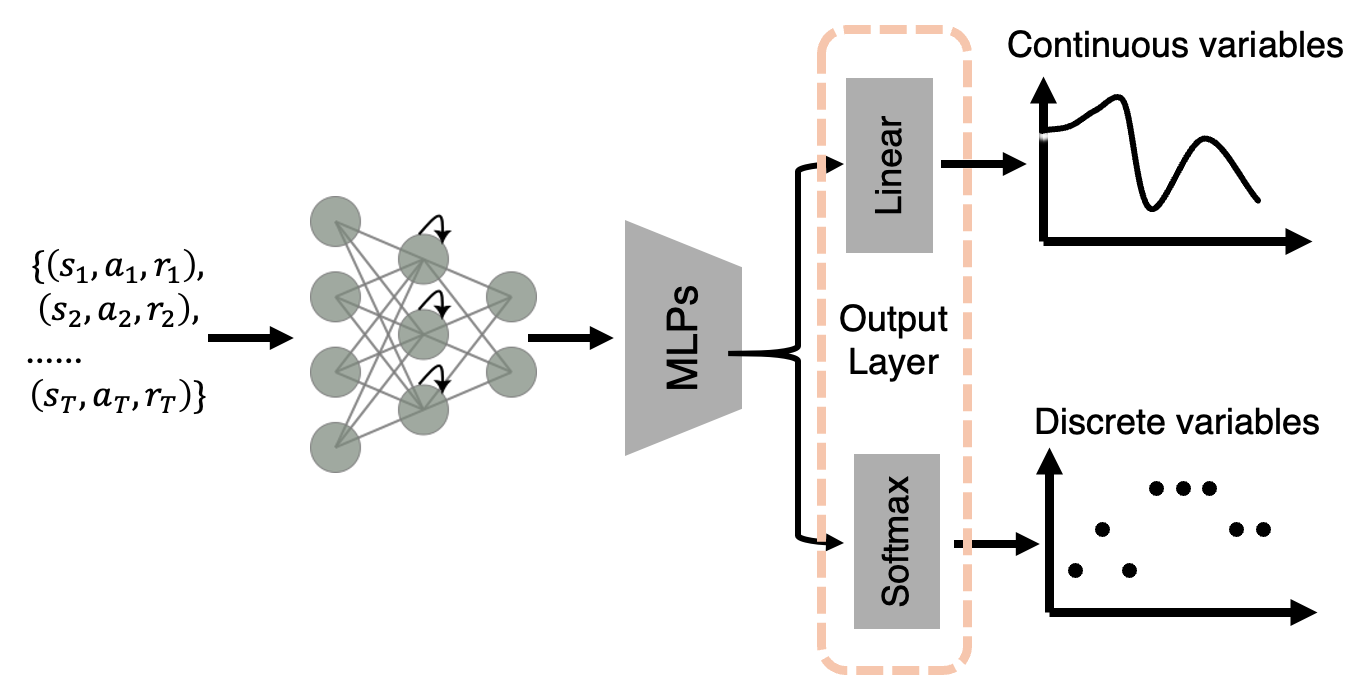}
    \caption{\textbf{Network structures}: The building blocks of LaseNet consist of one recurrent neural network (RNN) followed by two layers of Multilayer Perceptrons (MLPs). Depending on the goal of variable inference, we add either one or two different output layers to predict discrete and/or continuous latent variables; they receive the same embeddings from MLPs as input.}
    
    \label{fig:structures}
\end{figure}

Our proposed method relies on a two-phase approach, similar to standard ANN-based SBI methods: a training phase and an inference phase (Fig. \ref{fig:approach}). During the training phase, we first create a synthetic dataset by simulating the computational cognitive model of interest on the target experimental task. An artificial neural network (\emph{LaseNet}) is then trained on this synthetic dataset using model-simulated observable data $\mathbf{Y}$ as input and a series of model-derived latent variables $\mathbf{Z}$ as output. During the inference phase, the trained LaseNet takes the observable experimental data as input to infer a sequence of unobservable latent variables.
\paragraph{Architecture} LaseNet aims to learn a mapping between the observable variable space $Y$ and latent variable space $Z$. To learn this relationship, the structure of LaseNet is composed of two components: a bidirectional recurrent neural network (bi-RNN) layer \parencite{schuster1997bidirectional} followed by Multilayer Perceptrons (MLPs). The MLP layer has a pyramidal shape (Fig~\ref{fig:structures}), decreasing layer width from the largest to the output dimension.  After the bi-RNN layer, the number of units of each layer is half of the previous layer (See Table~\ref{table:hyper} for detailed specification).

The building blocks of bi-RNN are Gated Recurrent Units (GRU) \parencite{cho2014properties}. Bidirectionality enables the network to learn embeddings from both past and future history. A summary embedding $S$ is yielded by concatenating the past and future embeddings. We can represent the learned summary $S$ at each time point $y_t$ as:
\begin{equation} 
S(y_{t}) = \left \langle \overrightarrow{\psi}(\left \{ y  \right \}_{i=1}^{t}),\overleftarrow{\psi}(\left \{y  \right \}_{i=t}^{T}) \right \rangle 
\end{equation}
where function $\overrightarrow{\psi}$ (forward pass),  $\overleftarrow{\psi}$ (backward pass) transform a time series $(y_{1}, y_{2}...y_{T})$ to a lower-dimensional embedding space.  Following the bi-RNN, MLPs map the summary embeddings to the targeted latent variable space.  Let $\phi$ denote a universal function approximator. We can describe the estimated latent variable for each time point as $\tilde{z}_t = \phi(S(y_{t}))$.  When training LaseNet, the main objective is to find a set of neural network parameters (i.e., weights or biases) that minimizes the loss between true and estimated latent variables: $\mathcal{L}(\mathbf{Z};\mathbf{\tilde{Z}})$. 

\paragraph{Output layers} By changing the output layers, our architecture is adaptable to both continuous and discrete latent variables. For a continuous latent model variable space (e.g., Q-value), we used a linear activation with mean-squared error (MSE) loss. For a discrete latent model variable space (e.g., chosen strategy), we used a softmax activation function in the output layer, with a cross-entropy loss. To predict both types of latent spaces at once, we added two output layers for each type after the MLP layers (Fig~\ref{fig:structures}).

\paragraph{Details of neural networks Training} \label{para:training_details}
We used rectified linear unit (ReLU) as an activation function in all MLPs layers.  The trainings were run on Nvidia L4 GPUs, each equipped with 25 GB of memory, and required at most 30 minutes to complete (Appendix Table~\ref{table:training_time}). Network parameters were randomly initialized and optimized by Adam, with a learning rate of \(3*10^{-4}\). Each network was trained with at most 600 epochs and a batch size of 128. To avoid overfitting, we used 35 epochs as our early stopping criteria based on validation loss. Hyperparameters were fine-tuned using Bayesian optimization algorithms (Appendix Table~\ref{table:hyper}) with Bayesian optimization algorithms \parencite{bergstra2013making} applied to the validation set. We allocated 10\% of the training data as the validation set in fine-tuning.

\subsection{Benchmark settings}
\label{task}
\paragraph{Cognitive models and tasks}(Table~\ref{cognitive-models})  We evaluated LaseNet against five computational cognitive models and task environments. We first evaluated two tractable models: 4-parameter reinforcement learning model \textbf{(4-P RL)}\parencite{zou2022impulsivity}, and a meta reinforcement learning model with dynamic noise \textbf{(Meta RL)}\parencite{li2024dynamic}. We tested LaseNet's ability to infer both continuous and discrete variables from these cognitive models on a two-armed bandit learning task with probabilistic reversal \parencite{cools2002defining}. We then validated LaseNet in three likelihood intractable models: 
\begin{itemize}
    \item Hierarchical reinforcement learning \textbf{(HRL)} model performing a novel dynamic decision making task \parencite{rmus2023artificial}. In this task, participants observe three colored arrows, each pointing left or right (Fig. \ref{fig:approach}B). To earn rewards, participants choose a direction (left/right) based on the correct arrow, which changes unpredictably. This task is hierarchical because the correct choice (left/right) depends on a higher-level rule tied to arrow color. We assume that agents use a simple RL process to track arrow values and select a discrete arrow, before deciding on a side. Under these assumptions, the model is intractable, because the internal “arrow rule” selected by agents is unobservable and conditions on Q-value updates. It is one of the targets of latent variable estimation. 
    
    \item \textbf{Weber-imprecision} model (a variant of Bayesian generative model) \parencite{findling2021imprecise} performing a stimulus-action learning task in a volatile environment \parencite{collins2012reasoning}. In this task, participants learn to match three stimuli to four different actions. The volatile environment shifts between distinct sets of stimulus-action mappings with a probability of 3\%, requiring participants to continuously adapt to new stimulus-action associations. These mappings (or task sets) are latent states, with 24 possible mappings/latent states in total.
    
    \item 3-state \textbf{GLM-HMM} \parencite{ashwood2022mice} performing a standard two-choice perceptual decision-making task \parencite{international2021standardized}.
    
\end{itemize}

\begin{table}
  \caption{Summary of five computational cognitive models. a, r, and s denote actions, rewards, and stimuli, respectively. For example, 7-d s indicates that the stimulus input is coded as a 7-dimensional one-hot input vector.}
  \label{cognitive-models}
  \centering
  \begin{tabular}{lcccc} 
    \toprule
    \textbf{Name} & \textbf{\# of $\theta$} & \textbf{Input of LaseNet ($Y$)} & \textbf{Output of LaseNet ($Z$)} & \textbf{Tractable}  \\
    \midrule
    4-P RL  & 4  & \makecell{$\mathbb{R}^{2}$ \\ \textrm{[a, r]}} & \makecell{$\mathbb{R}^{1}$ \\ \textrm{Q-value}} &Yes  \\
    \midrule   
    Meta RL & 9 & \makecell{$\mathbb{R}^{2}$ \\ \textrm{[a, r]}}&  \makecell{($\mathbb{R}^{1}$ , $\mathbb{R}^{2}$) \\ \textrm{(Q-value, attentive state)}}& Yes  \\
    \midrule 
    HRL     & 2      & \makecell{$\mathbb{R}^{5}$ \\ \textrm{[a, r, 3-d s]}}& \makecell{($\mathbb{R}^{1}$ , $\mathbb{R}^{3}$) \\ \textrm{(Q-value, chosen cue)}} & No \\
    \midrule
    Weber-imprecision     & 3      & \makecell{$\mathbb{R}^{9}$ \\ \textrm{[a, r, 7-d s]}}& \makecell{($\mathbb{R}^{1}$ , $\mathbb{R}^{24}$) \\ \textrm{(distance, chosen mapping)}} & No \\    
    \midrule
    GLM-HMM & 21     & \makecell{$\mathbb{R}^{3}$ \\ \textrm{[a, r, 1-d s]}}& \makecell{$\mathbb{R}^{3}$ \\ \textrm{attentive state}} & No \\
    \bottomrule
  \end{tabular}
\end{table}

\paragraph{Dataset} For each LaseNet estimator, we simulated at most 9000 pairs of $(\textbf{Z},\textbf{Y})$ with 720 trials in each pair as training data (representing a standard cognitive task duration for a real biological agent).  We hold out 10$\%$ of training data as the validation set to fine-tune the hyperparameters. We simulated additional unseen 1000 pairs of  $(\textbf{Z},\textbf{Y})$ as testing data. 

\paragraph{Other estimators} As a comparison against LaseNet, we tested four commonly used likelihood-dependent estimators: MLE, MAP \parencite{wilson2019ten}, expectation–maximization (EM) \parencite{dempster1977maximum}, and sequential Monte Carlo (SMC) \parencite{doucet2000sequential, chopin2013smc2} on the same dataset. Two steps are required for researchers to recover latent variables with likelihood-dependent estimators. The first step is to find the best fitting model parameters $\theta$ given observable data \textbf{Y} by either maximizing a likelihood function $P(\mathbf{Y}\mid\theta)$ or posterior probability $P(\mathbf{Y}\mid\theta)\,\,P(\theta)$.  
In the second step, we used the best fitting parameters $\tilde{\theta}$ and observable data \textbf{Y} to derive the latent variables \textbf{Z}.  We then approximate latent variables from the targeted cognitive models \textbf{f} with $\tilde{\theta}$ and \textbf{Y} as input:
\begin{math}
\mathbf{\tilde{Z}} \approx  \mathbf{f}(\mathbf{Y};\tilde{\theta})
\end{math}. See Appendix~\ref{sup:likelihood-methods} for more detailed descriptions of the methods. 

\section{Synthetic Dataset Results}
\label{benchmark}

\subsection{Evaluation Metrics}
For all the metrics in this work, results were averaged across trials, where marker and error bars represent the mean and 2 standard deviations over all test samples.
\label{sup:metrics}
\paragraph{Root Mean-squared error} RMSE measures the difference between true ($Z_c$) and predicted continuous latent variables ($\hat{Z_c}$)(e.g., Q-values) from the estimators. RMSE across trials 1 to $T$ is defined as:
\begin{equation}
\text{RMSE}\,(\mathbf{Z_c},\mathbf{\hat{Z_c}})=  \sqrt{\frac{1}{T}\sum_{i=1}^{T}(z_i - \hat{z}_i)^2}    
\end{equation}
\paragraph{Negative log loss} NLL measures the predicted probability $\mathbf{p}$ from the estimators based on ground-truth discrete labels. We use NLL to quantify the performance of inferring discrete latent variables (e.g., attentive states and chosen cues). NLL across trials 1 to $T$ and a set of $M$ labels is defined as:
\begin{equation}
\text{Log Loss}\,(\mathbf{Z_d},\mathbf{P}) = -\frac{1}{T}\sum_{i=1}^{T}\sum_{j=1}^{M}z_{ij} \log(p_{ij})    
\end{equation}
\paragraph{Accuracy} We computed the balanced accuracy \parencite{brodersen2010balanced} that avoids inflated performance estimates on imbalanced datasets. It is the macro-average of recall scores per state. Thus for balanced datasets, the score is equal to accuracy. We used balanced accuracy to quantify the performance of inferring discrete latent variables, by taking the state with the highest probability as the predicted state. Balanced accuracy is defined as:
\begin{equation}
\text{Balanced Accuracy} = \frac{1}{2}\left(\frac{\text{TP}}{\text{TP} + \text{FN}} + \frac{\text{TN}}{\text{TN} + \text{FP}}\right)    
\end{equation}
where TP is true positive, TN is true negative, FP is false positive, and FN is false negative.

\subsection{Tractable models}
\label{tractable-model}
We tested LaseNet using synthetic datasets generated from two tractable models within the RL model family. Maximum Likelihood Estimation (MLE) was used as a benchmark for comparison.

\begin{figure}[htbp]
    \centering
    \includegraphics[width=0.9\linewidth]{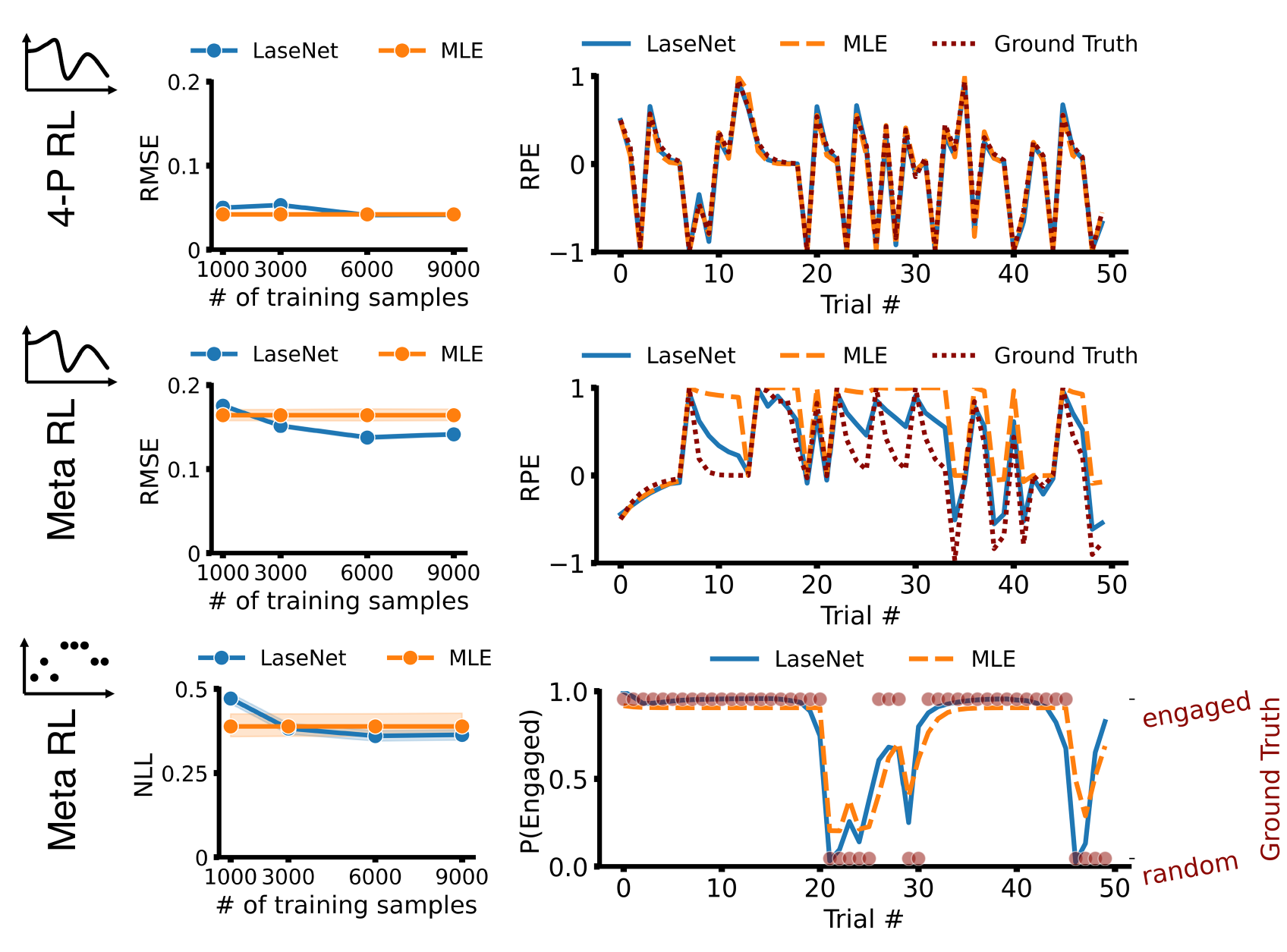}
    \caption{\textbf{Performance on synthetic dataset: tractable models}. The top row shows a strong agreement between LaseNet and MLE in 4P-RL model. The middle and bottom row present the results from the Meta RL model, showing that LaseNet identifies both continuous (Q-values) and discrete (engaged vs. random) latent variables with comparable precision but reduced variance (narrower shaded regions) compared to MLE. The right column displays example time series plots for one simulated participant from each model. Note that the reward prediction error (RPE) is obtained by subtracting estimated Q values from rewards.}
    \label{fig:tractable}
\end{figure}

\paragraph{4P-RL} The 4P-RL model is simulated in a 2-choice probabilistic reversal learning environment and follows the same Q-value update described in \nameref{preliminaries} with minor variants that bring it closer to human behavior \parencite{zou2022impulsivity}; the policy is a softmax over Q-values: $P(a_i) \propto \exp(\beta Q(a_i))$, with inverse temperature $\beta$ controlling noise in the policy. All further model and experiment details are available in Appendix \ref{sup:ccm}. Our target latent variable is the chosen Q-values, representing the subjective reward expectation at each trial. We found LaseNet reaches a similar average RMSE (\text{M} = 0.041) to a standard MLE-based approach (\text{M} = 0.042) (first row in Fig. \ref{fig:tractable}), showing our approach's capability to recover latent variables in a simple model and environment. We next tested it in a more complex cognitive model, where we can infer both continuous and discrete latent variables. 
\paragraph{Meta RL} Meta RL with dynamic noise model shares a similar Q-value update policy as 4-P RL. The major difference is that the model assumes a participant has two latent attentive states: engaged and random. The transition from one latent state to the other is controlled by a hidden Markov process \parencite{li2024dynamic}.  Two time-varying latent variables are inferred here: chosen Q-values (continuous latent space) and two attentive states (discrete latent space). We found that the RMSE of LaseNet (\text{M} = 0.141, \, \text{SD} = 0.038)  is slightly lower with less variance compared to MLE (\text{M} = 0.164, \, \text{SD} = 0.108)(last two rows in Fig. \ref{fig:tractable}), after training with 6k simulated participants. Moreover, in identifying attentive states, LaseNet had a slightly lower (better) NLL (\text{M} = 0.363) than MLE (\text{M} = 0.388). (last row in Fig. \ref{fig:tractable}), which may be due to parameter recovery issues with MLE (detailed in  Appendix~\ref{sup:mle-map}). 

\subsection{Intractable models}
\label{intractable}
In this section, we assessed LaseNet using three distinct intractable models: the HRL model (RL-based model), the Weber-imprecision model (Bayesian generative model), and the GLM-HMM (state-space model). For comparison, we used Sequential Monte Carlo (SMC) and Expectation-Maximization (EM), two widely used methods for approximating likelihoods and inferring time-varying latent variables in intractable cognitive computational models.

\begin{figure}[htbp]
    \centering
    \includegraphics[width=0.9\linewidth]{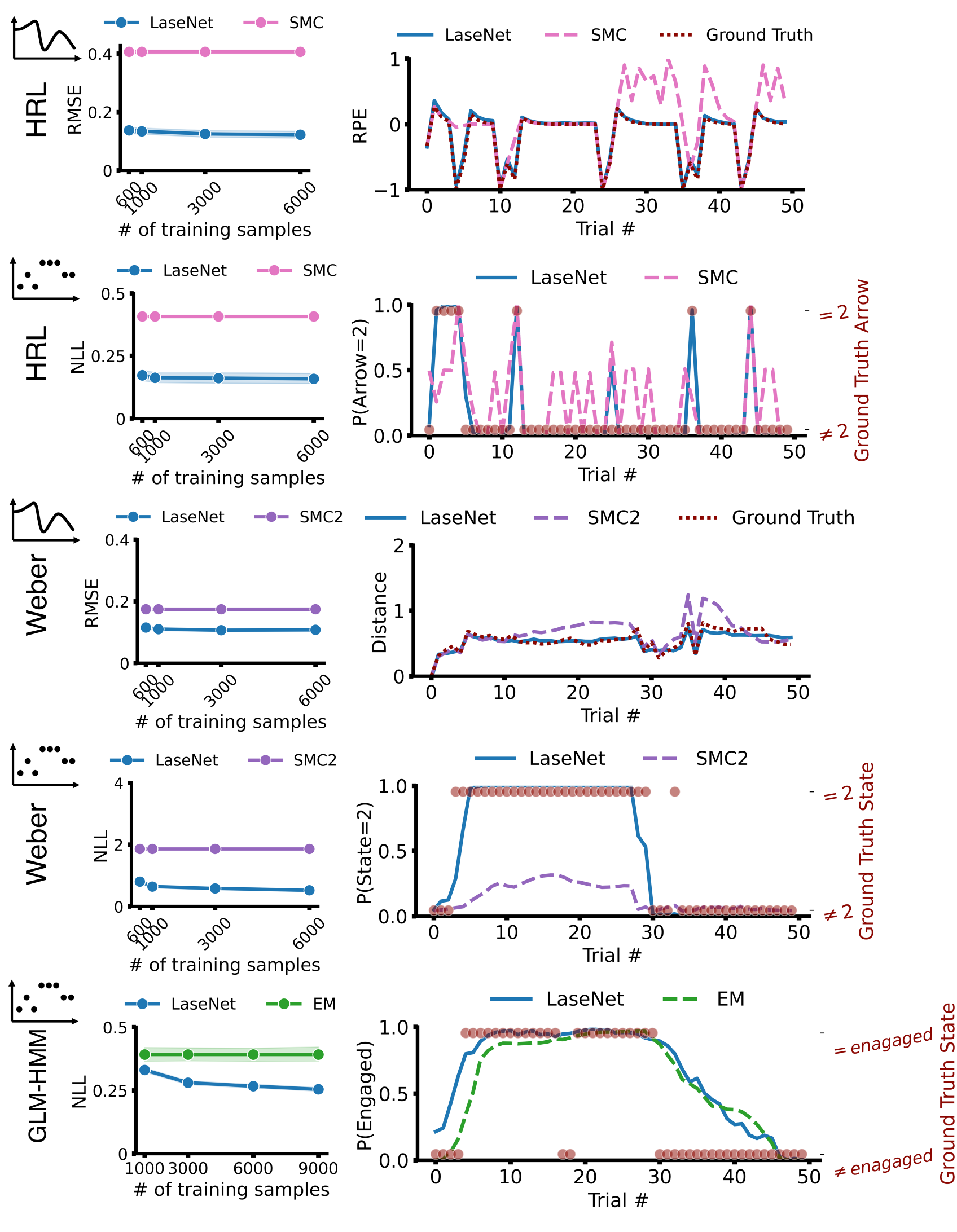}
    \caption{\textbf{Performance on synthetic dataset: intractable models}. The left column presents quantitative metrics across the three models, showing that LaseNet achieves lower (better) RMSE and NLL compared to other estimators. The right column shows trial-by-trial predictions from all estimators for an example agent. Overall, LaseNet demonstrates closer alignment with the ground truth.}
    \label{fig:intractable}
\end{figure}

\paragraph{HRL} The model assumes that participants track the value of each arrow, and choose between the arrows noisily:
\begin{math} \label{eq:hrl}
P(arrow) \propto \exp(\beta\;Q_{t}(arrow))
\end{math}. Since arrow choices are unobservable, the model’s likelihood is analytically intractable (See Appendix \ref{sup:hrl} for more details). The target latent variables include the selected Q-values (representing the value of the chosen arrow) and the chosen arrow. We used SMC, specifically particle filtering, to recover these time-varying latent variables. To isolate the performance of latent variable recovery, we assumed the model parameters were known when applying SMC, thereby bypassing the parameter recovery step. Note that this approximation should confer a benefit to the SMC approach compared to ours.
As shown in the top two rows of Fig.~\ref{fig:intractable}, LaseNet outperformed SMC. For Q-values identification, LaseNet achieved an average RMSE of 0.122 compared to 0.405 for SMC. In discrete latent cue identification (arrow selection), LaseNet reached 92.3\% accuracy, substantially surpassing SMC’s 71.5\%. Predicted cues were determined by selecting the cue with the highest probability from the estimators’ output.

\paragraph{Weber-imprecision} In a stimulus-action learning task, the model assumes that participants are more likely to change stimulus-action mapping (latent state) when the distance between posterior beliefs in consecutive trials is greater. Behavioral noise $\epsilon_t$ is modeled as a random variable uniformly distributed between 0 and $\mu + \lambda d_t$, where $d_t$ is the distance between posterior beliefs in trial $t-1$ and $t$. The parameters $\mu$ and $\lambda$ quantify Weber-imprecision.  The target latent variables are the posterior belief distance $d_t$ and the chosen state/mapping. We compared LasetNet with SMC2 \parencite{chopin2013smc2}, as used in \textcite{findling2021imprecise}. Consistent with the HRL benchmarking, we assumed known model parameters for SMC2. Results in Fig. \ref{fig:intractable} (middle two rows) demonstrate that LaseNet outperformed SMC2, with an average RMSE of 0.107 for LaseNet compared to 0.174 for SMC2. LaseNet achieved 54.7\% accuracy in latent state identification, while SMC2 reached 45\%, far above the chance level of 4\% (1/24). Additionally, LaseNet exhibited significantly lower NLL, indicating higher precision in predicting state probabilities (fourth row; right column in Fig.~\ref{fig:intractable}). 

\paragraph{GLM-HMM} A 3-state GLM-HMM is used to capture three attentive states: engaged, biased-left, and biased-right \parencite{ashwood2022mice} in a two-choice perceptual decision-making task. This model comprises three independent Bernoulli GLMs, each conditioned on the participant’s current discrete strategy state. Each GLM is characterized by a weight vector that defines how inputs are integrated into decision policies specific to that state.
We trained LaseNet to predict the time-varying attentive states and compared its performance to the approximate EM algorithm, as used in \textcite{ashwood2022mice}. We found that LaseNet achieved a lower average NLL of 0.255 compared to EM’s 0.392, indicating superior performance in predicting state probabilities (last row of Fig.~\ref{fig:intractable}). While the accuracy of both methods was comparable, LaseNet slightly outperformed EM, achieving 84.3\% accuracy versus EM’s 83.2\%.

\subsection{Prior misspecification}
\label{exp:prior}
We examined the impact of misspecified priors between LaseNet and the likelihood dependent methods in one tractable (4-P RL) and one intractable (GLM-HMM) model. Overall, LaseNets trained with the dataset generated from a less biased model prior resulted in a more robust performance. All LaseNets for 4P-RL were trained with 22k simulated participants and with 9k simulated participants for the GLM-HMM.

\begin{figure}
  \includegraphics[width=\linewidth]{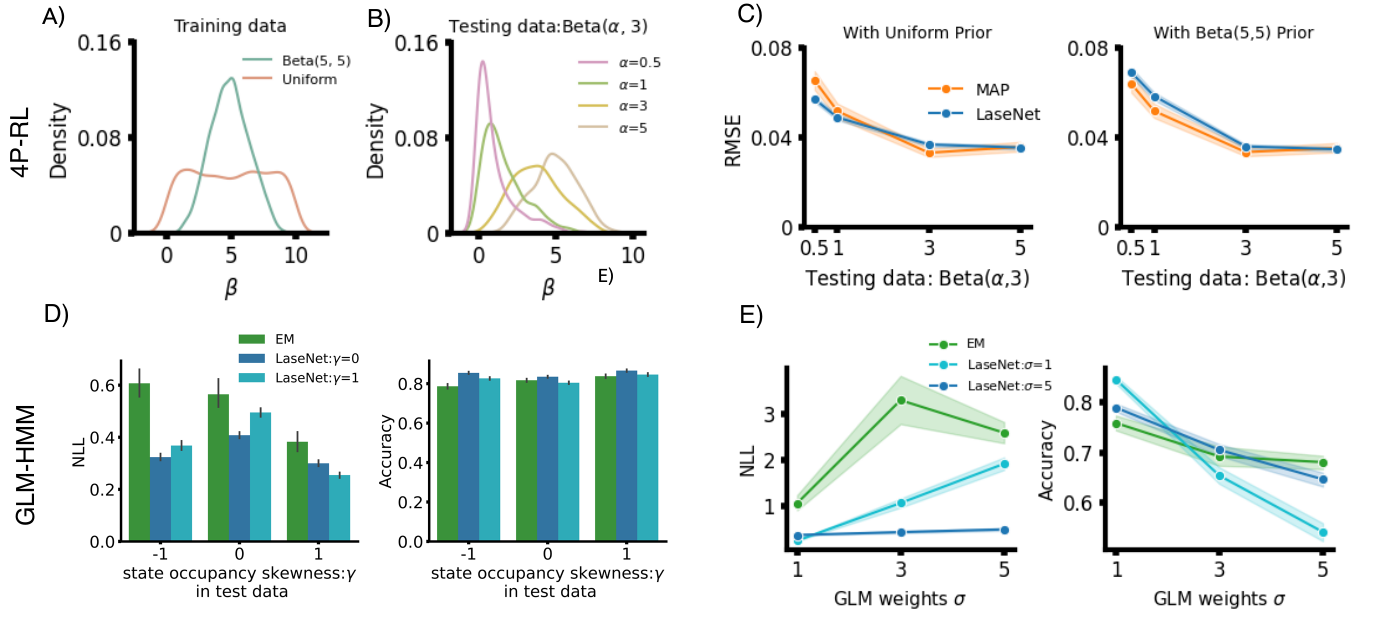}
  \caption{\textbf{Performance on misspecified priors}. Top row shows the impact of training with misspecified prior of model parameter $\beta$ in a 4-P RL model. (A) We compared two LaseNets trained with different prior distributions: a beta distribution and a uniform distribution. (B) Four test datasets generated from four different distributions of parameter $\beta$ are evaluated. (C) Training with a uniform distribution has more robust performance across different test datasets; Bottom row summarizes the impact of training with misspecified priors in a GLM-HMM model. (D) Three latent state distributions with positive (+1), no (0), and negative (-1) skewness $\gamma$ are evaluated. LaseNet trained with a uniform distribution (no skewness) outperforms both LaseNet trained with positive skewness and EM here. (E) Adjusting the $\sigma$ (noise level) in GLM weights in GLM-HMM models reveals that LaseNet is more robust when trained with noiser dataset (i.e., higher $\sigma$).}
  \label{fig:prior}  
\end{figure}

\paragraph{4P-RL} Softmax $\beta$ parameter in 4-P RL controls the randomness of a participant's action: higher beta results in a more deterministic actions. We evaluated two priors of $\beta$: a Beta ($\alpha$=5,$\beta$=5) prior (green line in Fig. \ref{fig:prior}A) and a uniform prior within an empirical range (orange line in Fig. \ref{fig:prior}A). We trained two LaseNet estimators with dataset generated from these two priors respectively. We tested the performance of LaseNet and MAP estimators with four different $\beta$ priors (red, light green, yellow and brown lines in Fig. \ref{fig:prior}B). We found that the LaseNet trained with a uniform prior had lower RMSE (0.044) across different $\beta$ priors (Fig. \ref{fig:prior}C).

\paragraph{GLM-HMM} Here, we changed the priors of the hidden state distribution (transitions matrix) and GLM weights, independently. In state distribution, we tested three skewness $\gamma$ levels: positive (1), negative (-1) and no (0) skewness. No skewness means that each state occupancy are equal (i.e.,uniform distribution).  We examined two LaseNet estimators: one is trained with positive skewness and the other is with no skewness in comparison with EM having a positive skewness prior.  We showed that LaseNet with equal states prior (no skewness) reaches the highest accuracy and lowest NLL among all (Fig. \ref{fig:prior}D). Furthermore, in changing the $\sigma$ in the GLM weights with a fixed mean, we found that LaseNet trained with a higher $\sigma$ (noiser) dataset is more robust (Fig. \ref{fig:prior}E). This suggests our approach is applicable even if there is no strong empirical priors.

\section{LaseNet Infers Latent Variable Sequence in Real Data}
\label{real-data}
\begin{figure}
  \includegraphics[width=1\linewidth]{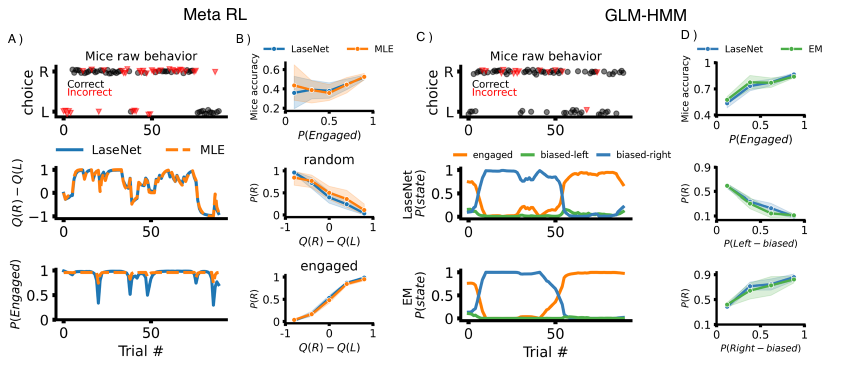}
  \caption{\textbf{Validation of LaseNet application on real mice data} (A) Top: raw behavioral data of one example mouse in the dynamic foraging dataset. Each dot represents a single trial; the y-axis indicates if the mouse went rightward or leftward on the trial. Middle: estimated Q-values difference. Bottom: latent engaged probability trial-by-trial in the example mouse. (B) LaseNet and MLE identify similar relationships between the mice response accuracy and probability of rightward choice given estimated latent variables. (C) Top: raw behavioral data of one mouse in IBL dataset. Middle and bottom rows: state probabilities from LaseNet and EM estimation, respectively, highlighting high agreement of certainty about the mouse’s internal state. (D) LaseNet and EM keep high consensus with the mice response accuracy and probability of rightward choice given different latent policy states.}
  \label{fig:real_data} 
\end{figure}

\paragraph{Meta RL inference in mice dynamic foraging dataset} The dynamic foraging dataset consists of 48 mice's data collected by \textcite{grossman2022serotonin}. Each mouse did a two-armed bandits task with dynamic reward schedules. We trained LaseNet to infer the latent attentive state (engaged vs random), Q-values of left and right actions. In comparison, we adopted MLE with an estimated likelihood function described by \textcite{li2024dynamic} as a benchmark. We found that in estimating Q-values, LaseNet had similar result as MLE. However, in attentive state identification, MLE tends to estimate state probability with high certainty (Fig. \ref{fig:real_data}A). In the behavioral analysis (Fig. \ref{fig:real_data}B), we found a similar trend between LaseNet and MLE: mice exhibit higher response accuracy when estimated engaged state probability is higher, and the inferred policy (probability of right choice $P(R)$ as a function of $Q(R)-Q(L)$) is consistent with the inferred latent states, validating the model assumptions.

\paragraph{GLM-HMM inference in mice decision making dataset} We used a mice decision making dataset published by the International Brain Laboratory (IBL) \parencite{international2021standardized}. The dataset consists of 37 mice performing a visual detection decision making task developed in \parencite{burgess2017high}. We extracted a time series of choice, reward and stimuli data from mice and fed into the trained LaseNet for inference. The LaseNet inferred time-varying probabilities for three HMM states (engaged, left-biased, right-biased). In comparison, we used EM fitting procedure described in \textcite{ashwood2022mice} as a benchmark. We obtained similar results in predicting state probabilities between LaseNet and EM (Fig. \ref{fig:real_data}C). Fig. \ref{fig:real_data}D shows a high agreement between LaseNet and EM, the mean absolute difference is 0.027 in mice accuracy, and 0.037 in right choice probability

\section{Discussion}
\label{discussion}
We proposed a novel method, LaseNet, for learning a mapping between observable data space and a targeted latent variable space using ANNs and synthetic data from cognitive model simulations. Unlike statistical estimators such as Maximum Likelihood Estimation (MLE) or Sequential Monte Carlo (SMC), LaseNet infers auxiliary latent variables without requiring likelihood computations. Additionally, our approach differs from neural point/posterior estimators: LaseNet focuses on identifying auxiliary latent variables within computational models instead of recovering model parameters. Our method does not require large data sets, works at the individual rather than group level, and can be applied to any cognitive model that is simulatable within a given experimental framework.

LaseNet offers several advantages over statistical estimators when applied to computational cognitive models. First, it performs comparably to statistical estimators in tractable models but is more robust in cases with many free model parameters (e.g., Meta RL). This robustness stems from its ability to bypass parameter recovery entirely, directly inferring auxiliary latent variables from observable data. Second, the flexible architecture of LaseNet can handle both continuous and discrete latent spaces, enabling a single network to simultaneously infer both types without performance degradation.

Furthermore, LaseNet excels in identifying latent variables within likelihood-intractable models, eliminating the need for custom and complex statistical estimators that approximate likelihoods, such as SMC or approximate EM algorithms. This makes LaseNet particularly advantageous for researchers working with simple intractable models, such as HRL, as it provides a generic framework for inference without requiring bespoke statistical methods.

When training with a dataset derived from less informative model parameter priors, LaseNet becomes more robust to different prior distributions in the test set, making it more likely to perform well on real participant data. To achieve high performance, the network does not require a large training sample (up to 22K simulated individual agents). It implies that, in a real-world experimental setting, researchers can use LaseNet to fit experimental data without using strong empirical model parameter priors and a high computational budget. Lastly, in comparison to other statistical estimators, LaseNet showed its generalizability to identify latent variables in a wide range of cognitive models (RL-based models without Markovian property and HMM-based models with Markovian property). 

\paragraph{Related Work}
Our framework is inspired by a wide range of studies that used ANN and SBI to overcome the challenges of likelihood computation in intractable models \parencite{cranmer2020frontier, radev2020bayesflow, rudi2022parameter}. Specifically, our work focuses on the approach of mapping data to variable point estimates using neural networks. Recent work \parencite{sainsbury2024likelihood} formalized the neural point estimation within a decision-theoretic framework. These neural point estimators date back to \parencite{chon1997linear} and have shown promising results in a variety of fields, including spatio-temporal forecasting \parencite{zammit2020deep}, spatial fields \parencite{gerber2020fast, lenzi2023neural} and time-series \parencite{rmus2023artificial}.

RNNs have been used widely in the processing of time-series data, a prevalent format in computational cognitive models.  While earlier studies have employed RNN-based models as cognitive models themselves \parencite{dezfouli2019disentangled, miller2024cognitive}, recent research has taken a different approach, leveraging RNNs as tools for parameter recovery and model identification in computational cognitive frameworks \parencite{rmus2023artificial, ger2024harnessing, ger2024using}. For example, \textcite{russek2024inverting} successfully showed the capability of RNNs to infer human preferences with simulated data.

Two recent studies have employed a similar workflow to identify time-varying variables using neural networks and simulation data. The first study adopts superstatistics framework to recover the dynamics of model parameters $\theta$ \parencite{schumacher2023neural}, which differs in objective from our approach, where we aim to infer auxiliary latent variables from the joint distribution $P(Z \mid \theta, Y)P(\theta \mid Y)$. The second study \parencite{ghosh2024sample} focuses solely on state space models (i.e., HMMs), but does not demonstrate the applications to model families lacking Markovian properties, such as RL models. Future work should compare LaseNet with the method proposed by \textcite{ghosh2024sample} directly on matched computational models and real dataset, as well as evaluate against various metrics for latent variable identification \parencite{lueckmann2021benchmarking}.

\paragraph{Limitations}
Though LaseNet is generalizable and flexible for various computational models, there are some limitations: First, similar to current statistical inference for cognitive modeling, LaseNet lacks uncertainty estimation. Consequently, LaseNet estimators are unable to detect out-of-distribution data. Uncertainty estimation is an important tool to detect the possibility that researchers may have model misspecification issues; i.e., that the real data does not come from the generative model family used for the LaseNet training data-set. In Appendix~\ref{sup:model-misspecified}, we showed examples of model-misspecification, where fitting a four-state GLM-HMM to a three-state GLM-HMM significantly reduces accuracy. Given the amortized nature of neural point estimators, boostrapping is frequently used to quantify uncertainty. However, boostrapping becomes computationally expensive in time-series prediction tasks. To mitigate this issue, one could perform a model identification first before training LaseNet or consider integrating with evidential learning \parencite{amini2020deep, sensoy2018evidential}; (See Appendix~\ref{sup:uq} for our experiments using evidential learning). This will be an important direction for future research. 

In addition, LaseNet does not infer underlying model parameters that are relevant to the targeted latent variables.  While it is less affected by models with many free parameters compared to standard estimators that rely on parameter recovery, understanding the relationship between model parameters and the resulting auxiliary latent variables may still be critical for researchers. Therefore, parameter recovery with other neural-based estimators \parencite{radev2020bayesflow, rmus2023artificial} should be used in conjunction with LaseNet to provide a more comprehensive analysis.

Finally, while we tested our method in a real data set, we could not compare the inferred model variable directly with ground truth, as this is by definition difficult to observe. We were able to demonstrate comparable qualitative results across methods, and thus indirectly validate our approach. Future work should consider using a dataset that includes different measurement proxies for the latent states (e.g., reaction time \parencite{botvinick2001conflict}, pupil size \parencite{laeng2012pupillometry}) and examine whether our proposed approach can outperform traditional statistical inference in real experimental settings. 

\section{Conclusion}
In this study, we present a new technique to extract latent cognitive model variables from observable behavior. LaseNet excels in handling computational models with intractable likelihoods, where traditional approaches are not applicable. The method is versatile, capable of identifying both discrete and continuous latent variables, and is generalizable across various computational models. Moreover, LaseNet is practical: it works with standard data sizes and can be applied at the individual participant level without high computational resources and technical expertise. Breaking down the barrier of intractable likelihood and recovering the latent dynamics of computational cognitive models will provide researchers with new insights into previously inaccessible dimensions in behavioral data. 

\section{Declarations}
\paragraph{Funding} This work was supported by the National Institutes of Health (NIH R21MH132974 to AGEC)
\paragraph{Conflicts of interest/Competing interests} The authors declare no conflicts of interest or competing interests.
\paragraph{Ethics approval} This paper contains only de-identified secondary data and is therefore not considered human subjects research.
\paragraph{Consent to participate/Consent for publication} Not applicable.
\paragraph{Availability of data and materials} The raw data studied in this paper are publicly available. The dynamic foraging dataset can be accessed at ~\url{https://datadryad.org/stash/dataset/doi:10.5061/dryad.cz8w9gj4s}. The IBL mice dataset can be accessed at ~\url{https://doi.org/10.6084/m9.figshare.11636748}. 
\paragraph{Code availability} We used Tensorflow for all neural networks. Code to reproduce results is available at ~\url{https://github.com/ti55987/lasenet}

\printbibliography

\newpage

\subfile{supplementary}

\end{document}

%% file: supplementary.tex
\appendix

An overview of the contents covered by the ensuing appendices:
\begin{enumerate}[label=\Alph*.]
    \item \nameref{sup:ccm}: including 4-P RL, Meta RL, HRL, Weber-imprecision, and GLM-HMM cognitive models used in this work.
    \item \nameref{sup:nn}: including network training, computational cost, likelihood-dependant methods, and real-data experimental details.
    \item \nameref{sup:robustness}: experiments on model misspecification and trial length variation.
    \item \nameref{sup:uq}: example integration with evidential deep learning.
\end{enumerate}

\section{Descriptions of Computational Cognitive Models}
\label{sup:ccm}
\subsection{Four-parameter reinforcement learning model}
On each trial t, the four-parameter reinforcement learning model (4P-RL) includes a stickiness parameter $\kappa$ which captures the tendency to repeat choice from the previous trial:
\begin{equation}
    P(a) \propto exp(\beta Q + \kappa \  \textrm{same}(a,a_{t-1}))
\end{equation}

Once the reward $r$ has been observed, the action values are updated based on \ref{eq:q-learning}. In addition, we adopt a counterfactual updating in this model, where the value of the non-chosen action also gets updated on each trial (\parencite{eckstein2022reinforcement}):
\noindent{}
\begin{equation}
\begin{gathered}
\delta_\text{unchosen} = (1-r) - Q_{t}(1-a) \\
Q_{t+1}(1-a) = Q_{t}(1-a) + \alpha\,\delta_\text{unchosen}
\end{gathered}
\end{equation}
\noindent{}
Instead of having one learning rate, the model differentiates between positive and negative feedback (\cite{niv2012neural}), by using different learning rates - $\alpha^+$ and $\alpha^-$ for updating action values after positive and negative outcomes respectively:
\begin{align*}
Q_{t+1}(a) &= \begin{cases}
Q_{t}(a) + \alpha^+\,\delta\ & \mbox{if } r\ > 0 \\
Q_{t}(a) + \alpha^-\,\delta\ & \mbox{if } r\ \leq 0 
\end{cases} 
\end{align*}
The 4P-RL model thus includes four free parameters: positive learning rate ($\alpha^+$), negative learning rate ($\alpha^-$), softmax beta ($\beta$) and stickiness ($\kappa$). As simulating data, all parameter priors are drawn from $\mathcal{U}(0,1)$, except for $\beta$, which is drawn from $\mathcal{U}(0,10)$.

\subsection{Meta reinforcement learning model with dynamic noise}
\label{sup:meta-rl}
The meta-learning model was first proposed by \textcite{grossman2022serotonin}. \textcite{li2024dynamic} extended the model with dynamic noise and shows  that the modified model fits the experimental data better. The model assumes that a participant has two latent states with time-varying transition probabilities ($T_0^1$, $T_1^0$), which denotes transitions from the random state to the engaged state and vice versa. In the random state, the decision policy is random and uniform. In the engaged state, on each trial t, a decision is sampled from choice probabilities obtained through a softmax function applied to the action values of the left and right actions with a bias. The probability of left choice is given by:
\begin{equation} 
P_t(l)=\frac{1}{1+exp(\beta \times \left ( Q_t(r)-Q_t(l) \right )+bias)}
\end{equation}
Regardless of the state that a participant is in, once the reward is observed, assuming the left action is chosen, Q-values are updated as follows:
\begin{equation}
Q_{t+1}(l)=Q_t(l)+\alpha_{t}\cdot \delta_t\cdot \left ( 1-E_t \right )
\end{equation}
where $\alpha_{t}$ is either $\alpha_{+}$ or $\alpha_{-}$ based on positive or negative outcomes, respectively, $E_t$ is an evolving estimate of expected uncertainty calculated from the history of absolute reward prediction errors (RPEs):
\begin{equation}
\begin{gathered}
E_{t+1} = E_t + \alpha_v\cdot v_t \\
v_t=\left | \delta_t \right |-E_t
\end{gathered}
\end{equation}
When RPE is negative, the negative learning rate is dynamically adjusted and lower-bounded by 0:
\begin{equation}
\alpha_{(-)_{t}} = max(0, \,\, \psi\cdot (v_t+\alpha_{(-)_{0}}) + (1-\psi)\cdot \alpha_{(-)_{t-1}})    
\end{equation}
Lastly, the unchosen action is forgotten:
\begin{equation}
Q_{t+1}(unchosen\,\,action) = \xi \cdot Q_{t}(unchosen\,\,action)    
\end{equation}
The model thus has 9 parameters: two transition probabilities ($T_0^1$, $T_1^0$), softmax beta ($\beta$), bias(for the right action), positive learning rate  ($\alpha^+$), baseline negative learning rate ($\alpha^-_0$), learning rate of RPE magnitude integration ($\alpha_v$), meta-learning rate for unexpected uncertainty ($\psi$), and forgetting rate ($\xi$). As simulating data, all parameter priors are drawn from $\mathcal{U}(0,1)$, except that $\beta$ is from $\mathcal{U}(0,20)$

Note that the likelihood is tractable in Meta RL, because the model assumes that the latent state only affects the policy to choose an action. In the random state, information is thus still being processed (e.g., Q-values updating given rewards) \parencite{li2024dynamic}. 

\subsection{Hierarchical reinforcement learning}
\label{sup:hrl}
The hierarchical reinforcement learning (HRL) model follows a similar updating policy as 4-P RL except that the model tracks the values of each arrow-following rule.  With N arrows, the complete probability of choosing each arrow is given by:

\begin{equation}
    P(arrow_i) = \frac{\exp(\beta\;Q_{t}(arrow_i)}{\sum_{i = 1}^{N} \exp(\beta\;Q_{t}(arrow_i))}
\end{equation}

Once the agent selects an arrow, it greedily chooses the direction based on the side (left or right) indicated by the arrow. For simplicity, different from 4-P RL, the model has no stickiness $\kappa$ and has a single shared learning rate for both positive or negative outcomes. The model includes only 2 free parameters: learning rate $\alpha$ $\sim \mathcal{U}(0.4, 0.7)$; and softmax beta $\beta$ $\sim \mathcal{U}(1, 10)$. 
\paragraph{HRL likelihood is intractable:} The likelihood in the HRL model can be formally written as such: %
\begin{align*}
  \mathcal{L}(\theta) &= \sum_{t=1}^{T} \log \probP(a_t \mid h_t, \overline{h}_{t-1}, \theta)  \\
  &= \sum_{t=1}^{T} \log \probP \Bigl( \sum_{c=1}^{C} \probP(a_t \mid h_t, \text{rule}_t = c; \theta) 
  \probP(\text{rule}_t = c \mid \{\overline{rule}_{t-1}\}; \overline{h}_{t-1}; \theta) \Bigr) \\
\end{align*}
where T is the number of trials, $a_t$ denotes the action a participant chose (left/right), $c\in \{1:C$\} denotes identity/color of the correct arrow, $\overline{h}_{t-1}$ corresponds to the observable task history encoding rewards, selected actions/sides, arrow directions (stimuli), while $\{\overline{rule}_{t-1}\}$ represents all possible unobservable the chosen arrows sequences in the past. This likelihood is computationally intractable due to the need to integrate over the uncertainty of what rule (which arrow) the participant followed in all of the past trials.  For example, if $C=3$ and $t=2$, there would be $C^T$ possible sequences (i.e., $3^2=9$ sequences of $\{(rule_1=1, rule_2=1), (rule_1=1, rule_2=2), (rule_1=2, rule_2=3)...(rule_1=3, rule_2=3)\}$). Thus, the integration exponentially increases with each trial, such that computing the likelihood requires summing over $C^T$ possible sequences of latent variables, making this impossible for any realistic value of $T$. 

\subsection{Weber-Imprecision}
\label{sup:weber}
The Weber-Imprecision model, a variant of the Bayesian generative model \parencite{findling2021imprecise}, assumes that participants form posterior beliefs B(t) about the correct mapping at each trial t.  Here, the term “latent state” refers to the mapping tracked by participants. To model action selection, we applied the softmax rule to the logarithm of these beliefs:
\begin{equation}
P_i(t) \sim \exp(\beta \cdot \ln B_i(t))
\end{equation}
where $P_i(t)$ denotes the probability of choosing action $i$ in trial $t$, $B_i(t)$ is the belief that action $i$ is the correct action in trial $t$ (calculated by marginalizing over latent state beliefs) and $\beta$ is the inverse temperature parameter. Unlike the sparse reward environment in \textcite{findling2021imprecise}, our setup delivered rewards $r_t$ for correct actions in every trial. The task involved three stimuli and four actions, resulting in 24 possible mappings.

In volatile environments, the Weber-Imprecision model adapts to changes in the correct mapping by adding noise with a probability $\epsilon_t$ described in \nameref{intractable}.  When the current latent state changes between trial $t-1$ and $t$, the new latent state $z_t$ is drawn from the set of potential latent states $\{1, \dots, K\}$ according to a multinomial distribution with probabilities $\boldsymbol{\gamma} = \{\gamma_1, \dots, \gamma_K\}$, where K is 24. 
\begin{equation}
z_t \mid z_t \neq z_{t-1} \sim \text{Multinomial}(\boldsymbol{\gamma})
\end{equation}
with $\sum_{k=1}^{K} \gamma^k = 1$ and excluding the preceding state. The priors probability over $\gamma$ is uninformative and follows a flat Dirichlet distribution. The model includes three free parameters: $\mu$ $\sim \mathcal{N}(0.05, 0.003)$, $\lambda$ $\sim \mathcal{N}(1.22, 0.06)$, and $\beta$ $\sim \mathcal{N}(5, 0.023)$.

\subsection{Hidden Markov Models with Bernoulli generalized linear model observation}
\label{sup:glmhmm}
We use a framework based on HMMs with Bernoulli generalized linear model (GLM) observations to analyze decision-making behaviors in mice \parencite{ashwood2022mice}. The resulting GLM-HMM, also known as an input-output HMM, supports an arbitrary number of states that can persist over a large number of trials and have different dependencies on the stimulus and other covariates. A GLM-HMM consists of two basic components: an HMM that governs the distribution across latent states and a set of state-specific GLMs, specifying the strategy employed in each state. For a GLM-HMM with K latent states, the HMM has a $K\times K$ transition matrix $A$ specifying the probability of state transition,
\begin{equation}
P(z_t = k \mid z_{t-1} = j) = A_{jk}    
\end{equation}
where $A_{jk}$ denotes the transition from state $j$ to state $k$, $z_{t-1}$ and $z_t$ indicate the latent state at trials $t-1$ and $t$, respectively. To represent the state-dependent mapping from inputs to decisions, the GLM-HMM comprises K independent GLM weights, each defined by a vector $w$ indicating how inputs are integrated in that particular state (described in Eq~\ref{eq:glm}). We use 4-dimensional $w$ for inputs: stimuli, bias, previous choice, and win-stay/lose-switch.  With a three-state GLM-HMM, there are 21 free parameters: $\mathbf{\theta} \equiv \left \{ A, {w}_{k=1}^{k=3}  \right \}$, which $A$ is a $3\times 3$ transition matrix and $w_k \in \mathbb{R}^4$ is a GLM weight vector for state $k$.

Similar to HRL, both the Weber-imprecision, and GLM-HMM require computing likelihoods by summing over all possible $K$ latent states in the $T$ trials. This results in $K^T$  terms as marginalizing all possible paths.

\section{Further Experimental Methods Details}
\label{sup:nn}
\subsection{ANN training specification}
\label{sup:laset-spec}
\begin{table}
  \centering  
  \caption{Hyperparameters selection.}
  \label{table:hyper}
  \begin{tabular}{lll}
    \toprule
    \cmidrule(r){1-2}
    Hyperparameter    & Sweep range      \\
    \midrule
    \# of units in RNN layer  &  $\mathcal{U}(36, 326)$ \\
    dropout rate in RNN layer &   $\mathcal{U}(0.05, 0.25)$ \\
    dropout rate in 1st MLP layer  & $\mathcal{U}(0.01, 0.1)$ &  \\
    dropout rate in 2nd MLP layer & $\mathcal{U}(0.01, 0.05)$  \\
    learning rate & 0.003 or 0.0003 \\
    \bottomrule   
  \end{tabular}
\end{table}

\begin{table}[H]
\centering

\caption{Training time (\(\text{mean} \pm \text{std. dev.}\) in minutes) of LaseNet on two likelihood-intractable models that predict both continuous and discrete latent variables. Each LaseNet was trained with a learning rate of $3e-4$ and a batch size of 256 on the L4 GPU provided by Google Colab. }  
\label{table:training_time}  

\begin{tabular}{m{3cm}|m{1cm}m{1cm}m{1cm}m{1cm}|m{1cm}m{1cm}m{1cm}m{1cm}}
\toprule
\makecell{\textbf{Models}} & \multicolumn{4}{c|}{\textbf{HRL}} & \multicolumn{4}{c}{\textbf{Weber-imprecision}} \\ 
\midrule
\makecell{\textbf{\# of Trials}} & \multicolumn{4}{c|}{\textbf{720}} & \multicolumn{4}{c}{\textbf{300}} \\ 
\midrule
\makecell{\textbf{\# of Samples}} & \makecell{\textbf{600}} & \makecell{\textbf{1000}} & \makecell{\textbf{3000}} & \makecell{\textbf{6000}} & \makecell{\textbf{600}} & \makecell{\textbf{1000}} & \makecell{\textbf{3000}} & \makecell{\textbf{6000}} \\ 
\midrule
Mean & 4.25 & 6.28 & 17.36 & 28.44 & 2.72 & 3.51 & 7.84 & 12.02 \\ 
Std. Dev & 0.015 & 0.011 & 0.017 & 2.2 & 0.21 & 0.3 & 0.82 & 0.97 \\ 
\bottomrule
\end{tabular}

\end{table}

\subsection{Details of Likelihood-dependent Methods}
\label{sup:likelihood-methods}
In this appendix, we described in details of using the likelihood-dependent method to infer latent variables. The fitting process consists of two steps: identifying the best-fitting parameters and then inferring the latent variables with best-fitting parameters. Here, we used coefficient of determination $R^2$ and Pearson correlation coefficients $r$ to measure the performance of parameter recovery. $R^2$ represents the proportion of variance in true parameters that can be explained by a linear regression between true and predicted parameters. Best possible $R^2$ score is 1.  Pearson correlation coefficients $r$ represent the strength of a linear association between true and predicted parameters. Since a positive correlation is desired, the best possible $r$ score is 1.

\subsubsection{Maximum likelihood and Maximum a posteriori estimation}
\label{sup:mle-map}
\begin{figure}
\includegraphics[width=0.9\linewidth]{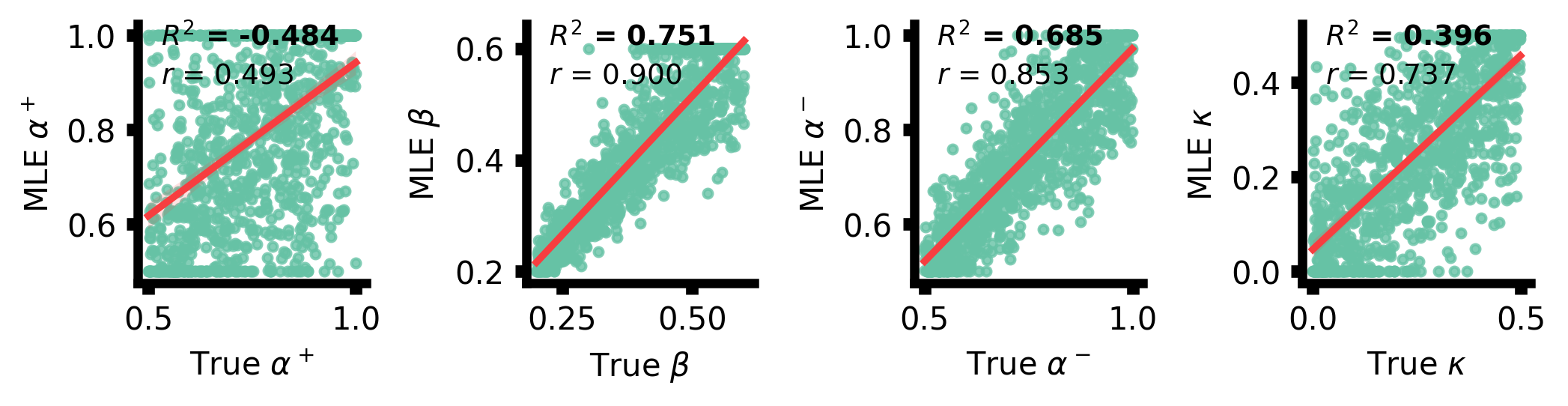}
\centering   
  \caption{\textbf{4-P RL parameters recovered by MLE} $R^2$ is a R-squared score. $r$ corresponds to Pearson correlation coefficient, red line represents a least squares regression line. High correlation score shows that MLE recovers well in this simple RL model.}
\label{fig:4prl_parameter_recovery} 
\end{figure}

\begin{figure}
\includegraphics[width=0.7\linewidth]{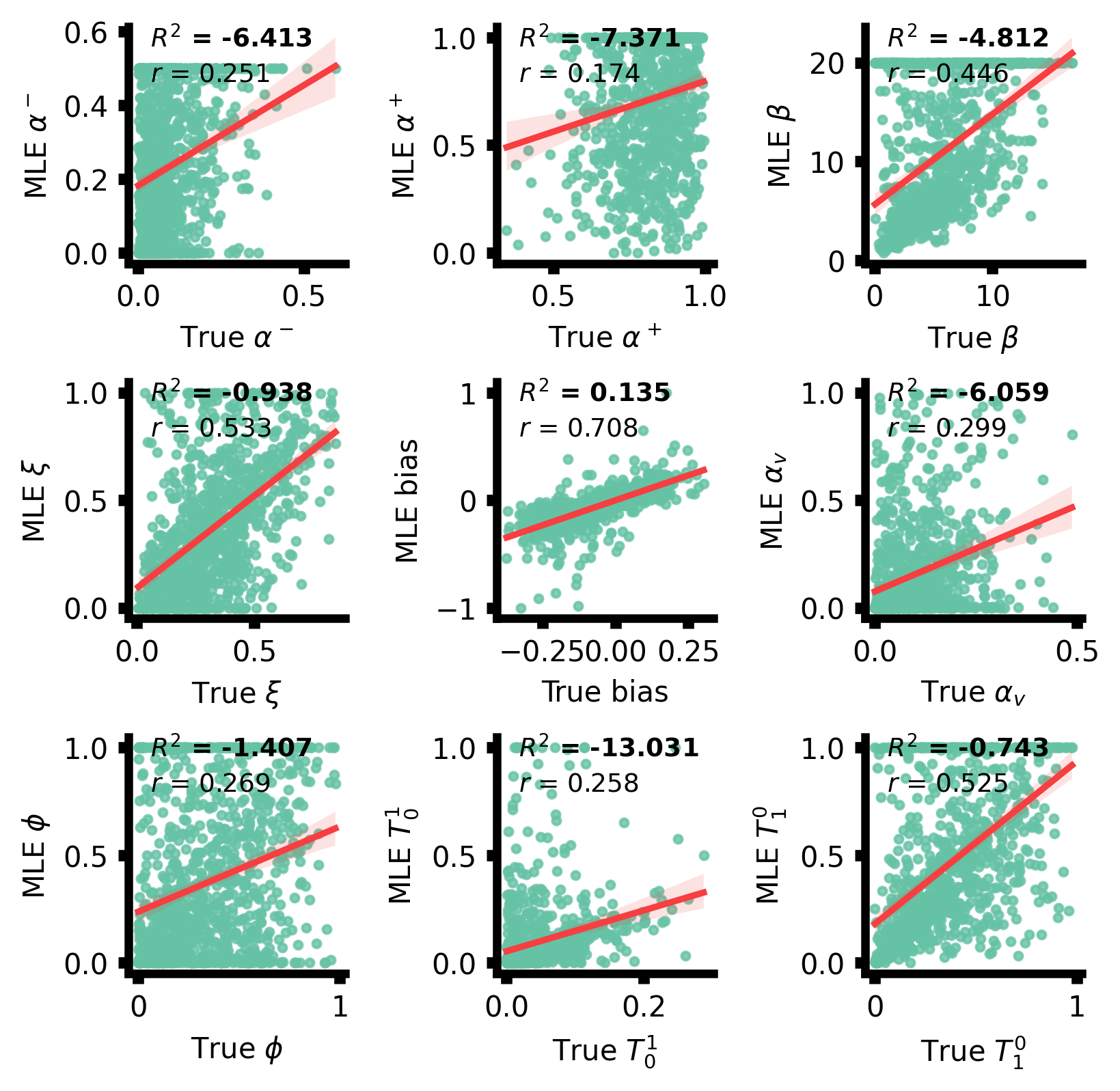}
\centering   
  \caption{\textbf{Meta RL parameters recovered by MLE} Unlike Fig.~\ref{fig:4prl_parameter_recovery}, MLE shows fairly limited parameter recovery in a highly parameterized model, which may impair MLE-dependent latent variable estimation.}
\label{fig:df_parameter_recovery} 
\end{figure}

Maximum likelihood estimation (MLE) leverages probability theory and estimation of likelihood $P(Y\mid\theta)$ of the data given the model parameters and assumptions (\cite{myung2003tutorial}). The best-fitting parameters are determined by maximizing the log-likelihood of the data: 
\begin{equation} 
\theta_{MLE} = \argmax_{\theta} P(Y \mid \theta) = \argmax_{\theta} \sum_{i} \text{log}\, P(y_{i}\mid\theta)
\end{equation}
To find the parameters via MLE, we used the Sequential Quadratic Programming (SQP) method provided by MATLAB fmincon. SQP allows for solving the optimization problem with constraints. The constraints we imposed on are the empirical ranges of each model parameter (e.g., [0, 1] for parameters' search space). Fig~\ref{fig:4prl_parameter_recovery} and Fig~\ref{fig:df_parameter_recovery} show the recovered parameters in experiments with tractable models. With the recovered parameters and the observable data $Y$, we then derived the latent variables such as Q-values by running through the RL functions.

Maximum a posteriori estimation (MAP) relies on a similar principle, with an addition of a prior $P(\theta)$ to maximize the log posterior: 
\begin{equation} 
\theta_{MAP} = \argmax_{\theta} \sum_{i} \,[\,\text{log}\, P(y_{i}\mid\theta) + \text{log}\,P(\theta)\,]
\end{equation}
We employed MAP in prior misspecification experiments~\ref{exp:prior} for fitting data generated from a 4-P RL model.  Specifically, we tested different $\beta$ parameters prior (Fig~\ref{fig:prior}, keeping other priors with uniform distributions. Same as MLE, we used the SQP algorithm in MATLAB and derived latent variables with the recovered parameters and the observable data.

\subsubsection{Sequential Monte Carlo}
\label{sup:smc}

Sequential Monte Carlo (SMC) methods \parencite{gordon1993novel} approximate probability distributions with weighted particles that evolve as new observations arrive, enabling inference in nonlinear, non-Gaussian dynamic systems. The particle weights reflect likelihood, providing an estimate of the posterior distribution and its uncertainty \parencite{samejima2003estimating}. In our study, the posterior distribution is $P(Z \mid Y_{1:t}, \theta)$, where $Z$ denotes targeted latent variables (particles), $Y_{1:t}$ denotes the observable variables from trial 1 to t, and $\theta$ are known model parameters.  In the HRL model, we applied particle filtering (PF) algorithms to recover the latent arrows/cues. Specifically, the PF consists of procedures:
\begin{enumerate}
    \item Propagate: particles are propagated based on the probability of choosing each arrow (Appendix ~\ref{sup:hrl}).
    \item Update: particle weights are updated based on the agent’s actions and the associated Q values. For instance, if the agent selects the unique arrow, the particle weight corresponding to that arrow is set to 1, while all others are assigned a weight of 0. Conversely, if the agent chooses a non-unique arrow, the weights are adjusted proportionally to the Q values.
    \item Resample: if the $\text{Effective Sample Size (ESS)} = 1/\sum w_i^2$ \parencite{liu1995blind} is less than half of the number of particles, particles are resampled based on the weights.
\end{enumerate}
We implemented the algorithm by iterating step 1 to step 3 across 720 trials and using a total number of 1000 particles. Each particle is initialized randomly from a uniform distribution.

In benchmarking the Weber-imprecision model, we employed the same method used in \textcite{findling2021imprecise}: $\text{SMC}^2$, an algorithm extends from iterated importance sampling \parencite{chopin2002sequential} and particle Markov chain Monte Carlo method \parencite{doucet2000sequential}. We used a total number of $51200$ particles over 300 trials.

\subsubsection{Expectation-Maximization}
\label{method:em}
\begin{figure}
\includegraphics[width=1\linewidth]{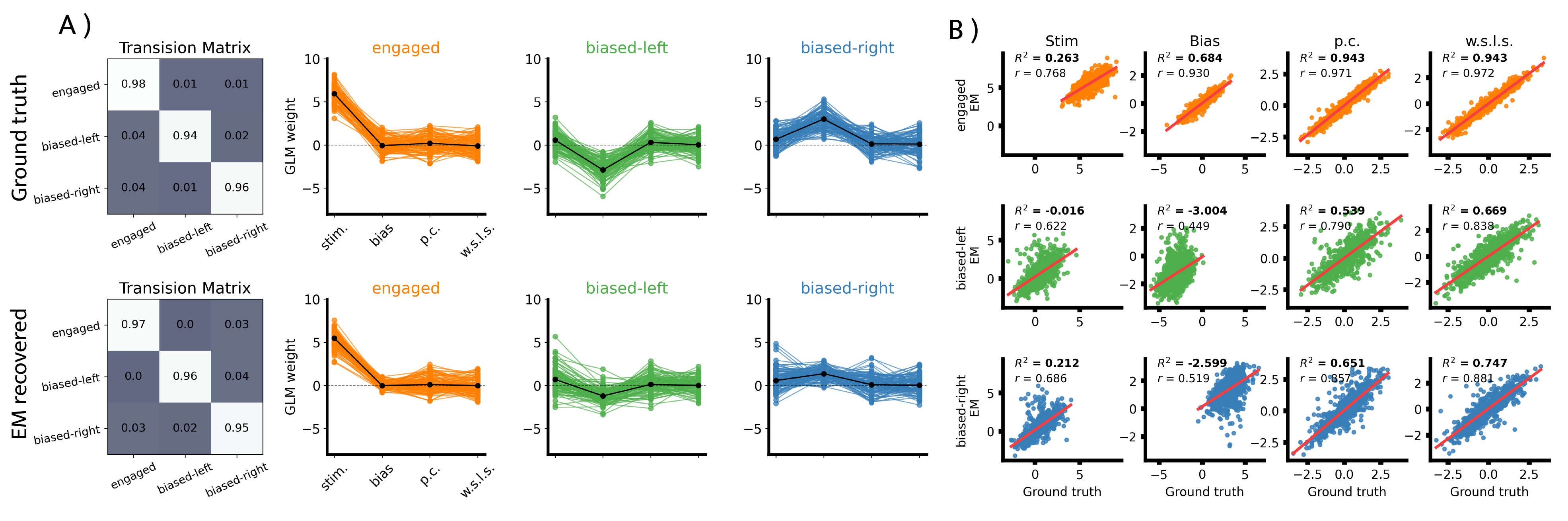}
\centering   
  \caption{\textbf{3 states GLM-HMM recovered by EM} (A) Top row: simulated ground truth of transition matrix and GLM-weights in three states: engaged, biased-left, and biased-right. Bottom row: the well-recovered parameters from EM: high engaged state in the transition matrix, high stimulus GLM weight in the engaged state, negative bias weight in the left-biased state, and vise versa (GLM takes left as negative; right as positive value). (B) Each row represents GLM weights in three different states and each column represents four weighted features: stimuli, bias, previous choice, win-stay/loss-switch. In the GLM-HMM case, we showed that good recovery in parameters doesn't guarantee high precision in latent variables estimation compared to LaseNet (Fig.~\ref{fig:intractable}).}
\label{fig:glmhmm_parameter_recovery} 
\end{figure}

Expectation-maximization (EM) algorithm is an iterative method for determining the maximum likelihood or maximum a posteriori estimates of parameters in statistical models based on unobserved latent variables $Z$. The log-likelihood to be maximized is thus given by marginalizing out latent variables $Z$:
\begin{equation}
\label{eq:em-log}
\text{log}\,P(Y\mid\theta) = \sum_{z} \text{log}\,P(Y, Z \mid \theta)
\end{equation}

Nonetheless, this is intractable in GLM-HMM model as described in~\ref{proof:glmhmm}. Hence, following the same approach in \textcite{ashwood2022mice}, we maximize the complete log likelihood instead. During the E-step of the EM algorithm, we compute the expected complete data log-likelihood (ECLL), which is a lower bound on Eq.~\ref{eq:em-log}.  We can derive the lower bound as:
\begin{align*}
\text{log}\,P(Y\mid\theta) &= \text{log}\,\sum_{z} P(Y, Z \mid \theta)  \\
&= \text{log}\,\sum_{z} P(Z \mid Y, \theta)\,\,\frac{ P(Y, Z \mid \theta)}{P(Z \mid Y, \theta)} \\
&= \text{log}\left(\mathbb{E}_{P(Z\mid Y, \theta)}\left [\frac{ P(Y, Z \mid \theta)}{P(Z \mid Y, \theta)} \right ]\right)\\
&\geq \mathbb{E}_{P(Z\mid Y, \theta)}\,\,\text{log}\left [\frac{ P(Y, Z \mid \theta)}{P(Z\mid Y, \theta)} \right ]
\end{align*}
Then, during the ‘maximization’ or M-step of the algorithm, we maximize the ECLL with respect to the model parameters $\theta$. To understand how this iterative algorithm converges to the desired log-likelihood, please see \textcite{wu1983convergence} for the actual proof. 

Fig~\ref{fig:glmhmm_parameter_recovery} shows the recovered transition matrix and GLM weights in a three-state GLM-HMM model after applying EM. We then infer the most probable latent states by running the E-step with the best-fitting parameters and the input data.

 \begin{figure}
\includegraphics[width=1\linewidth]{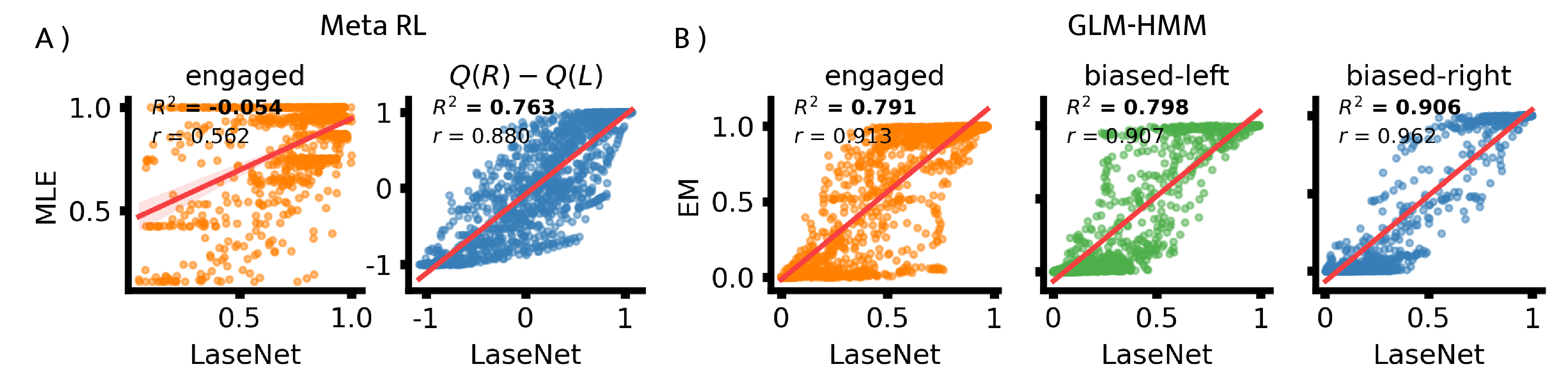}
\centering   
  \caption{\textbf{Correlation between LaseNet and likelihood-dependent methods in real data} (A) In Meta RL, MLE and LaseNet has similar estimated engaged probability and strong correlation in estimated Q-values difference. (B) In GLM-HMM, EM and LaseNet shows high positive correlation in predicting state probability.}
\label{fig:real_data_corr} 
\end{figure}

\subsection{Infer Latent Variables in Mice Dataset}
\label{sup:details-real-data}
In this section, we outlined the training details of LaseNet for real mice data.
\paragraph{Dynamic foraging dataset} We first simulated 9000 participants from Meta RL model described in Appendix~\ref{sup:meta-rl}; each simulated participant has 720 trials.  The parameters of Meta RL model were drawn from empirical distribution given by ~\parencite{li2024dynamic}.  We then trained LaseNet with the simulated participants; LaseNet takes a time series of rewards and actions as input (Table~\ref{cognitive-models}), and outputs three time-varying latent variables: attentive states, Q-values of leftward action, and Q-values of rightward action.  Compared to MLE, we found higher correlation in Q-values difference estimation, than latent state probability estimation (Fig.~\ref{fig:real_data_corr}A). 


\paragraph{IBL dataset} We simulated 6000 participants from GLM-HMM model described in ~\ref{sup:glmhmm}; each simulated participant has 500 trials. We used an equal transition probability between states. GLM weights were drawn from empirical distribution given by~\parencite{ashwood2022mice}. We trained LaseNet with simulated participants. LaseNet takes a time series of rewards, actions, and stimuli as input (Table~\ref{cognitive-models}), and outputs one latent variable with three possible states:engaged, left-biased and right-biased. Fig.~\ref{fig:real_data_corr}B shows high correlation between EM and LaseNet in inferring latent state probabilities.


\section{Additional Robustness Tests}
\label{sup:robustness}
\begin{figure}
\includegraphics[width=1\linewidth]{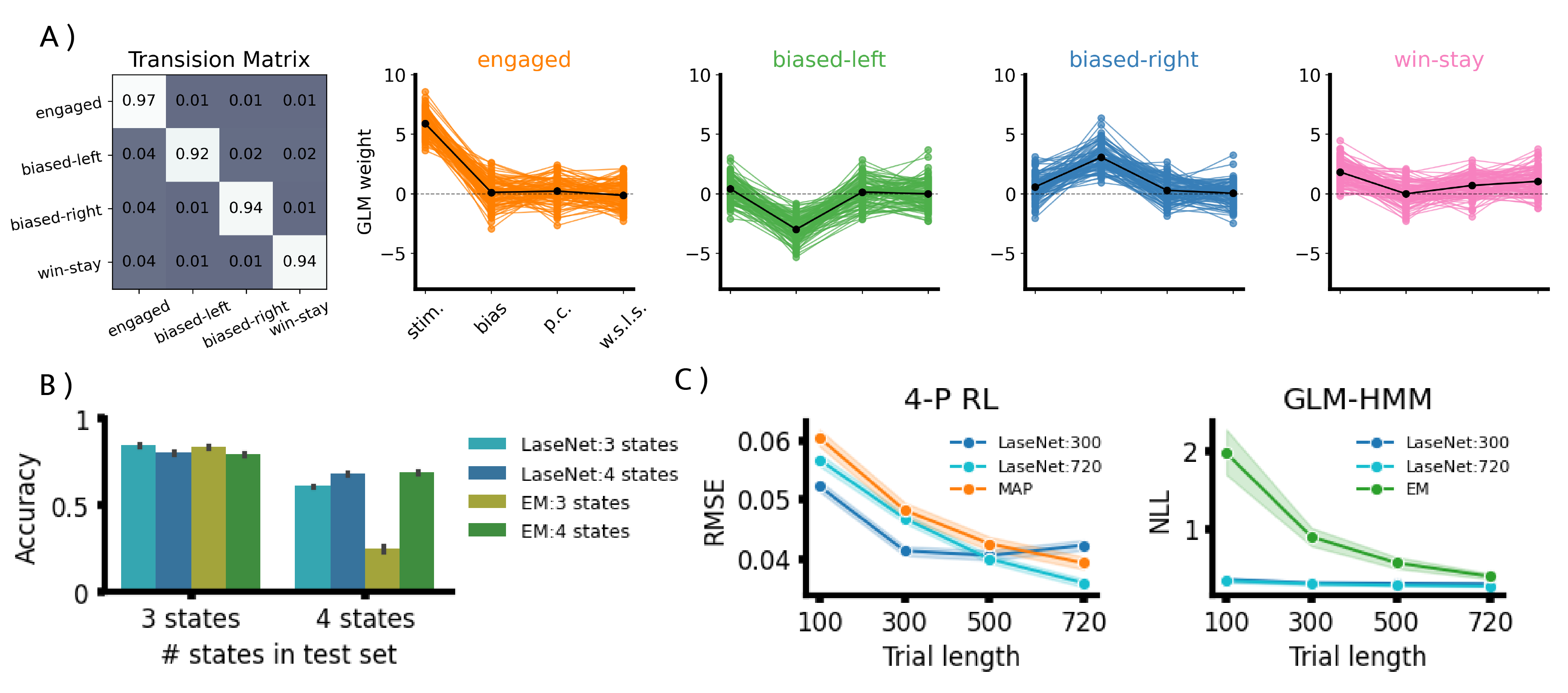}
\centering   
  \caption{\textbf{Robustness Tests} (A) A four state GLM-HMM with additional win-stay state compared to three-state GLM-HMM in Fig.~\ref{fig:glmhmm_parameter_recovery}. A win-stay has a relatively high win-stay, loss-switch weight in GLM. (B) Accuracy drops when fitting with a wrong model prior in both EM and LaseNet. A model identification is thus needed before inferring the latent states. (C) In both 4-P RL and GLM-HMM, LaseNet trained with longer trials exhibits more robust results across different trial lengths.}
\label{fig:robustness} 
\end{figure}

\subsection{Model misspecification}
\label{sup:model-misspecified}
We tested model misspecification by fitting data generated from a 3-states GLM-HMM to estimators with a 4-states prior, and vice versa. For LaseNet estimators, we trained two neural networks with data generated from 3 and 4 states priors, respectively. Data generated from a 3-states GLM-HMM has the same model prior as previous experiments (Fig~\ref{fig:glmhmm_parameter_recovery}A), while data from a 4-states GLM-HMM has the prior shown in Fig~\ref{fig:robustness}A.  We added a "win-stay" state for a 4-states GLM-HMM based on empirical results in \parencite{ashwood2022mice}.  Each LaseNet estimators was trained with 3000 simulated samples. We generated additional 500 simulated samples for each state priors as test sets. All samples have 720 trials. We compared our trained LaseNet estimators with EM having either 3 or 4 states priors.  We found that both EM and LaseNet have lower accuracy when fitting with misspecified models (Fig~\ref{fig:robustness}B). Hence, model identification should be performed before inferring latent states.

\subsection{Trial length variation}
We examined LaseNet in inferring latent states with different trial's lengths, because in real experiments, researchers commonly collect varying trial's lengths. We tested two cognitive models, 4-P RL and GLH-HMM, with four different trial's lengths:100, 300, 500, and 720. In comparison to likelihood-dependent methods, we trained two LaseNet estimators with data generated from 300 and 720 trials, respectively. Each LaseNet estimators was trained with 22000 simulated samples for 4-P RL and with 9000 simulated samples for GLM-HMM. For testing, we generated additional 500 simulated samples for each trial length. Fig~\ref{fig:robustness}B shows that training with 720 trials reaches higher precision consistently across all trial lengths compared to likelihood-dependent methods. Note that EM is susceptible to a short trial length because we can only approximate likelihood for GLM-HMM as described in~\ref{method:em}; shorter trial length yields less data points for approximation.

\section{Uncertainty Quantification}
\label{sup:uq}
\begin{figure}
\includegraphics[width=1\linewidth]{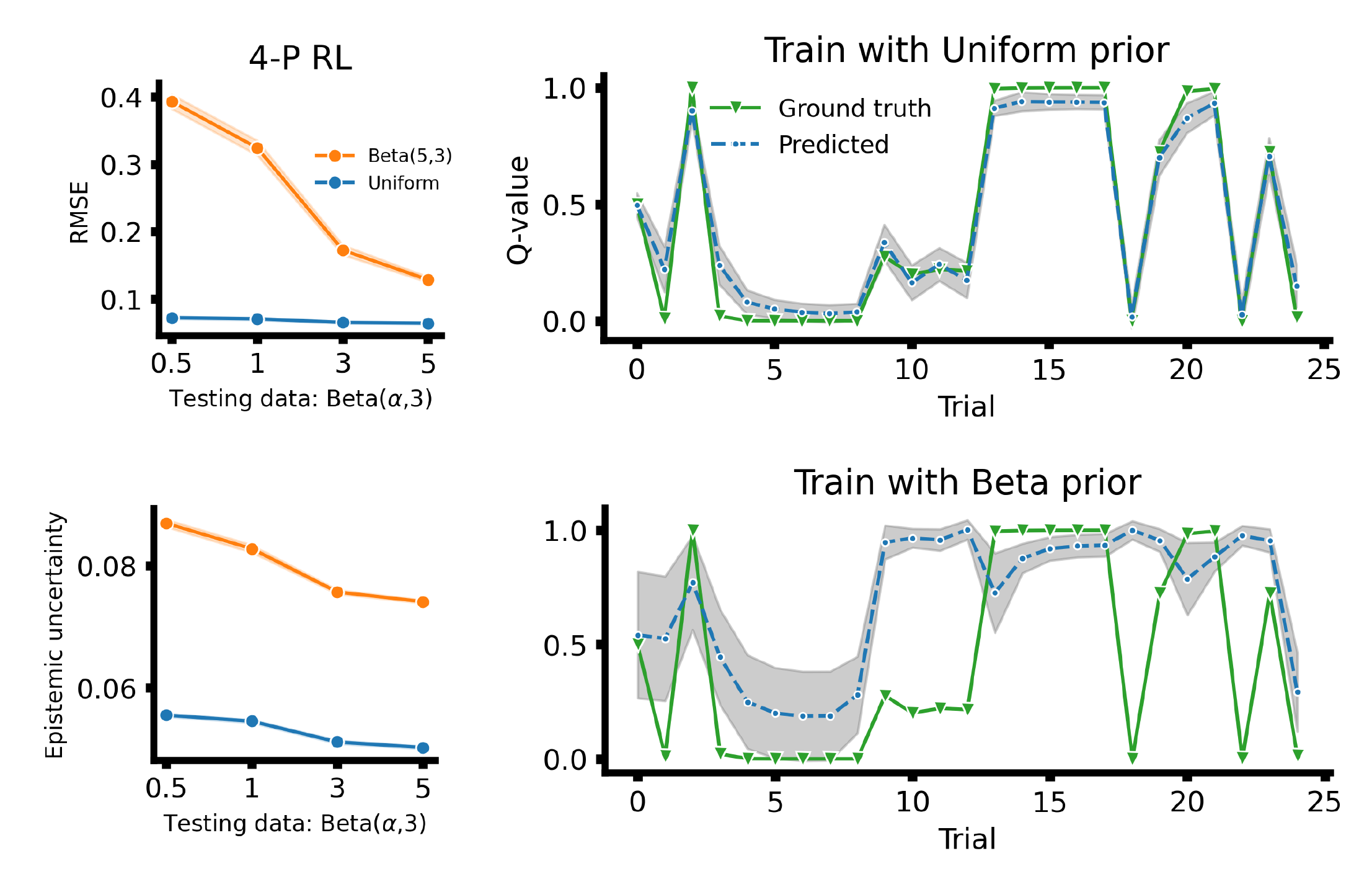}
\centering   
  \caption{\textbf{Uncertainty Quantification in 4-P RL} Two LaseNet estimators: one is trained with a uniform prior and the other is trained with a beta prior (Fig~\ref{fig:prior}A). LaseNet trained with a uniform prior has lower RMSE and uncertainty as testing with unseen data generated from different distributions. In right columns, grey area shows $\pm 1$ standard deviation.}
\label{fig:uq} 
\end{figure}
In this appendix, we explored the potential of integrating with evidential deep learning \parencite{amini2020deep, sensoy2018evidential}. Evidential deep learning aims to train a neural network to learn a higher-order, evidential distribution and then output the hyperparameters of the evidential distribution. Here, we adopted the loss functions proposed in \parencite{amini2020deep} to quantify the uncertainty for inferring Q-values in a 4-P RL model. The network structure is the same as Fig~\ref{fig:nn} except that the objective function is maximizing and regularizing evidence.  To measure the effect of evidential deep learning, we compared two LaseNet estimators trained with different $\beta$ model priors:uniform and beta distribution, illustrated in Fig~\ref{fig:prior}A.  Each LaseNet estimator was trained using 3000 simulated samples, each consisting of 500 trials. Fig~\ref{fig:uq} shows that the LaseNet estimator trained with a uniform prior exhibits lower RMSE and uncertainty. Higher uncertainty also corresponds to higher RMSE.  Our implementation is extended from:~\url{https://github.com/aamini/evidential-deep-learning}
